\documentclass[conference]{IEEEtran}

\usepackage{cite}
\usepackage{amsmath,amssymb,amsfonts}
\usepackage{algorithmic}
\usepackage{algorithm}
\usepackage{graphicx}
\usepackage{textcomp}
\usepackage{xcolor}
\def\BibTeX{{\rm B\kern-.05em{\sc i\kern-.025em b}\kern-.08em
    T\kern-.1667em\lower.7ex\hbox{E}\kern-.125emX}}

\usepackage{booktabs}
\usepackage{multirow}
\usepackage{array}
\usepackage{hyperref}

\newcommand{\ACROLETTER}[1]{\underline{\text{#1}}}

\newcommand\methodname{\textcolor{black}{\textsc{SPEAR-MM}}}
\newcommand\methodnamelongform{\textcolor{black}{\textsc{\ACROLETTER{S}elective \ACROLETTER{P}arameter \ACROLETTER{E}valuation \ACROLETTER{A}nd \ACROLETTER{R}estoration via \ACROLETTER{M}odel \ACROLETTER{M}erging (SPEAR-MM})}}

\begin{document}

\title{SPEAR-MM: Selective Parameter Evaluation and Restoration via Model Merging for Efficient Financial LLM Adaptation}

\author{\IEEEauthorblockN{Berkcan Kapusuzoglu, Supriyo Chakraborty, Renkun Ni, Stephen Rawls and Sambit Sahu}
\IEEEauthorblockA{\textit{Capital One} \\
McLean, VA 22012, USA \\
berkcan.kapusuzoglu@capitalone.com}
}

\maketitle
\thispagestyle{plain}
\pagestyle{plain}

\begin{abstract}
Large language models (LLMs) adapted to financial domains often suffer from catastrophic forgetting of general reasoning capabilities essential for customer interactions and complex financial analysis. We introduce \methodnamelongform, a practical framework that preserves critical capabilities while enabling domain adaptation. Our method approximates layer-wise impact on external benchmarks through post-hoc analysis, then selectively freezes or restores transformer layers via spherical interpolation merging. Applied to LLaMA-3.1-8B for financial tasks, SPEAR-MM achieves 91.2\% retention of general capabilities versus 69.7\% for standard continual pretraining, while maintaining 94\% of domain adaptation gains. The approach provides interpretable trade-off control and reduces computational costs by 90\% crucial for resource-constrained financial institutions. 
\end{abstract}

\begin{IEEEkeywords}
Large Language Models, Continual Learning, Financial AI, Catastrophic Forgetting, Model Merging, Parameter Efficiency
\end{IEEEkeywords}

\section{Introduction}

Financial institutions increasingly require domain-specific language models that can understand regulatory documents, analyze market data, and provide accurate customer support while maintaining broad reasoning capabilities for complex financial scenarios. However, continual pretraining (CPT) and supervised fine-tuning (SFT) on proprietary financial corpora typically cause catastrophic forgetting, where models lose general reasoning skills measured by benchmarks like GSM8K (mathematical reasoning) and ARC (logical reasoning) that are essential for financial analysis \cite{sun2023comprehensive}.

Traditional continual learning approaches face significant challenges in financial settings. Rehearsal methods require storing sensitive data, violating privacy constraints. Parameter-efficient methods like LoRA \cite{hu2021lora} limit adaptation capacity. Regularization techniques like Elastic Weight Consolidation (EWC) \cite{kirkpatrick2017overcoming} require careful hyperparameter tuning and substantial computational overhead. Most critically, these methods provide little interpretable control over the forgetting-adaptation trade-off, making them unsuitable for regulated financial environments where model behavior must be predictable.

Recent advances in model merging \cite{matena2022merging,wortsman2022modelsoupsaveragingweights} and layer analysis suggest a different approach. The Spectrum tool \cite{hartford2024spectrumtargetedtrainingsignal} demonstrates that signal-to-noise ratios vary dramatically across parameters, while layer-wise analysis reveals functional specialization in transformers. We leverage these insights to develop a post-hoc approximation method that identifies which layers are critical for preserving general capabilities.

Our contributions include:
\begin{itemize}
\item \textbf{Layer Impact Approximation}: A computationally efficient method to estimate each layer's contribution to general benchmark performance without expensive retraining.
\item \textbf{Selective Preservation Framework}: A practical pipeline combining signal-to-noise ratio (SNR) analysis, layer-wise change metrics, and spherical interpolation merging.
\item \textbf{Financial Domain Validation}: Comprehensive experiments on LLaMA-3.1-8B showing preserved general reasoning with strong financial domain adaptation.
\end{itemize}

\section{Methodology}
\label{sec:method}

\subsection{Problem Formulation}
\label{sec:problem}

Let $M_0$ denote a pretrained foundation model (e.g., LLM) with broad general capabilities. Let $D_{prop}$ denote a proprietary corpus of internal financial documents used to adapt the model via continued pretraining (CPT) or supervised fine-tuning (SFT), yielding a domain-specialized model $M_{prop}$. Let $B_{ext}$ denote external public benchmarks that measure general capabilities.

While $M_{prop}$ typically improves performance on internal tasks, it often suffers degradation on $B_{ext}$ due to specialization drift or catastrophic forgetting. Repeated mixtures of CPT + freezing + LoRA tuning can alleviate this issue but require expensive retraining cycles and sensitive hyperparameter choices.

Our objective is to construct a merged model $M_{merged}$ that:

\begin{enumerate}
\item Preserves domain knowledge acquired in $M_{prop}$,
\item Retains general competencies from $M_0$, 
\item Avoids additional full training cycles,
\item Provides explicit controllability over the specialization--generalization tradeoff.
\end{enumerate}

We frame this as post-hoc parameter selection and interpolation: identify parameters in $M_{prop}$ that must be preserved to retain financial-domain capabilities, and selectively restore parameters from $M_0$ where specialization causes forgetting.

\subsection{Parameter Importance via Multi-Metric Scoring}

We quantify parameter importance by comparing $M_0$ and $M_{prop}$ at a per-layer and per-parameter level. Let $w_0$ and $w_{prop}$ denote corresponding parameter tensors in the base and adapted models.

We use two complementary metrics:
\begin{enumerate}
    
\item{SNR-Weighted Change Intensity (SWCI)} 
Measures relative parameter displacement normalized by parameter scale and weighted by spectral signal-to-noise ratio (SNR):

\begin{equation}
\text{SWCI}(p) = \frac{\|w_{prop} - w_0\|_F}{\|w_0\|_F + \epsilon} \cdot \text{SNR}(p),
\end{equation}

where SNR is estimated via eigenvalue spread of the parameter covariance. SWCI identifies large, reliable parameter movements that likely encode domain-specific knowledge.

\item{Singular Value Drop Ratio (SVDR)} 
Captures structural functional shifts via singular value spectra:

\begin{equation}
\text{SVDR}(p) = 
\frac{\sum_{i=1}^k [\sigma_i(w_0) - \sigma_i(w_{prop})]}
{\sum_{i=1}^k \sigma_i(w_0) + \epsilon},
\end{equation}

where $\sigma_i(\cdot)$ are top-$k$ singular values. SVDR measures change in information-carrying rank structure, revealing deeper functional relationships beyond raw magnitude.

\item{Combined Score: SPEAR-MM}

We define a normalized weighted fusion:

\begin{equation}
\text{SPEAR-MM}(p) = \alpha \cdot \tilde{\text{SWCI}}(p) + \beta \cdot \tilde{\text{SVDR}}(p)
\end{equation}

where $\alpha, \beta$ are hyperparameters. This yields a robust importance signal: SWCI captures magnitude-driven adaptation; SVDR captures structural adaptation.

Parameters are grouped by transformer submodules (MLP, attention Q/K/V/O projections) and ranked within each group to respect architectural roles.

\end{enumerate}

\subsection{Selective Restoration via Spherical Interpolation (SLERP)}

Given parameter rankings, we apply controlled restoration by interpolating between the adapted and base weights for the top-ranked fraction of parameters per component:

\[
W_l^{final} = \text{SLERP}(W_l^{prop}, W_l^0, t_l)
\]

where $t_l \in [0, 1]$ is a merge coefficient. SLERP ensures smooth transitions and norm preservation in weight space, outperforming linear interpolation for directional stability.

We evaluate three restoration policies:

\begin{itemize}
\item \textbf{Conservative}: restores fewer parameters (40\% MLP, 40\% Attention)
\item \textbf{Balanced}: moderate restoration (50\% MLP, 60\% Attention)
\item \textbf{Aggressive}: maximal general capability retention (60\% MLP, 95\% Attention)
\end{itemize}

These ratios encode the empirical observation that attention layers more strongly impact general reasoning, while MLP layers encode domain-specific representations.

\subsection{Algorithm}
\label{sec:algorithm}

\begin{algorithm}[h]
\caption{SPEAR-MM Model Merging Pipeline}
\label{alg:spearmm}
\begin{algorithmic}[1]
\REQUIRE Pretrained model $M_0$, domain-adapted model $M_{prop}$, preservation ratios
\STATE Compute spectral SNR scores for each weight matrix
\STATE Compute SWCI and SVDR for all trainable parameters
\STATE Compute $\text{SPEAR-MM}(p) = \alpha \tilde{\text{SWCI}} + \beta \tilde{\text{SVDR}}$
\STATE Group parameters by transformer component (MLP, Q/K/V/O projections)
\STATE Rank parameters within each group by SPEAR-MM score
\IF{deterministic configuration}
  \STATE Select top-$k$ parameters per group per policy (Conservative/Balanced/Aggressive)
\ELSE
  \STATE Perform Bayesian search over top-$k$ ratios
  \FOR{each candidate configuration}
      \STATE Apply SLERP-based restoration
      \STATE Evaluate on held-out $D_{prop}$ and $B_{ext}$
  \ENDFOR
\ENDIF
\STATE Output best $M_{merged}$ according to validation trade-off
\end{algorithmic}
\end{algorithm}

\subsection{Pipeline Overview}

\begin{figure}[t]
\centering
\includegraphics[width=0.48\textwidth]{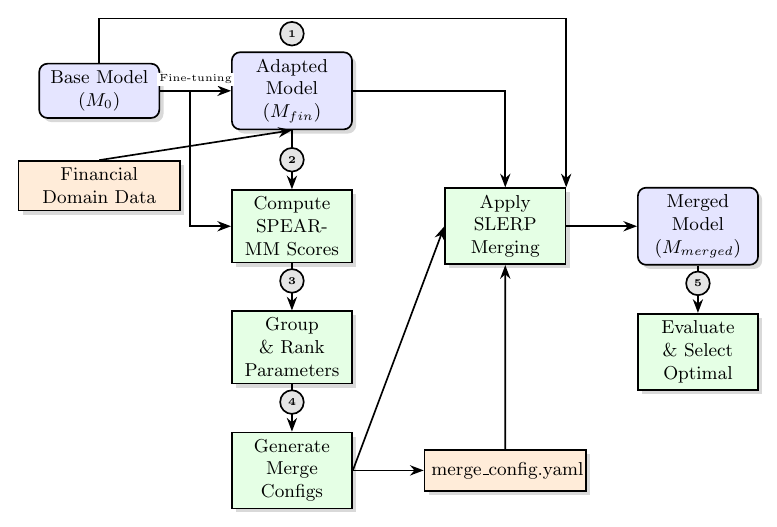}
\caption{\textbf{The $\methodname$ Pipeline:} (1) A base model is fine-tuned on financial data to create an adapted model. (2) Layer-wise $\methodname$ scores are computed using SNR and change metrics from both models. (3) Parameters are grouped and ranked by their score. (4) A merge configuration is generated to create the final model, which is then (5) evaluated on validation sets to select the optimal configuration.}

\label{fig:pipeline}
\end{figure}

Figure~\ref{fig:pipeline} illustrates the workflow: fine-tune base model, score parameter importance, rank and preserve parameters, SLERP-merge, evaluate, and select configuration.

\section{Experiments}

\subsection{Experimental Setup}

\textbf{Model}: LLaMA-3.1-8B with CPT on 500B tokens of proprietary dataset of internal business documents and optional SFT on 250K instruction pairs.

\textbf{Benchmarks}: ARC-Challenge (reasoning)~\cite{clark2018thinksolvedquestionanswering}, GSM8K (mathematics)~\cite{cobbe2021training}, MMLU/MMLU-Pro (knowledge)~\cite{wang2024mmluprorobustchallengingmultitask}, GPQA (advanced reasoning)~\cite{rein2023gpqagraduatelevelgoogleproofqa}, IFEval (instruction following)~\cite{zhou2023instructionfollowingevaluationlargelanguage}, MATH-HARD (complex mathematics)~\cite{hendrycks2021measuringmathematicalproblemsolving}.
Financial tasks: CloseQA (a factual task with $\sim$1,000 samples), OpenQA (a reasoning task with $\sim$1,000 samples), and a summarization task ($\sim$600 samples compared against human-annotated golden truths).

\textbf{Baselines}: Full CPT, CPT+SFT, CPT+LoRA (r=8), random freezing (50\%), top-8 layer freezing.

\subsection{Results}

Table \ref{tab:external_benchmarks} shows that full CPT causes catastrophic degradation: 69.5\% GSM8K retention, 53.1\% IFEval, 33.8\% MATH-HARD. Adding SFT partially recovers performance (82.0\% average), while LoRA improves further (85.6\%). SPEAR-MM dramatically outperforms all baselines: aggressive configuration achieves 91.2\% average retention, which is 21.5 points above full CPT. Mathematical reasoning shows particularly strong preservation: 97.5\% versus 69.5\%, critical for financial compliance.

\begin{table*}[htbp]
\centering
\caption{External Benchmark Retention (\% of baseline). Full CPT causes severe degradation especially in mathematical reasoning and instruction following. SPEAR-MM significantly mitigates forgetting.}
\label{tab:external_benchmarks}
\begin{tabular}{lccccccc|c}
\toprule
\textbf{Method} & \textbf{ARC-C} & \textbf{GSM8K} & \textbf{MMLU} & \textbf{MMLU-PRO} & \textbf{GPQA} & \textbf{IFEval} & \textbf{MATH-HARD}  & \textbf{Avg.} \\
\midrule
Base Model & 100.0 & 100.0 & 100.0 & 100.0 & 100.0 & 100.0 & 100.0 & 100.0 \\
Full CPT & 91.5 & 69.5 & 92.3 & 63.8 & 84.2 & 53.1 & 33.8 & 69.7 \\
CPT + SFT & 99.5 & 88.1 & 93.1 & 78.8 & 91.8 & 72.2 & 50.3 & 82.0 \\
CPT + LoRA & 99.8 & 90.9 & 92.4 & 82.1 & 94.3 & 84.4 & 55.3 & 85.6 \\
Random Freeze & 98.9 & 90.5 & 93.9 & 79.9 & 91.1 & 74.6 & 35.2 & 80.6 \\
Top-8 Freeze & 99.6 & 89.5 & 94.7 & 82.4 & 94.9 & 85.7 & 55.3 & 86.1 \\
\midrule
\textbf{SPEAR-MM (Conservative)} & \textbf{99.4} & \textbf{89.6} & \textbf{95.3} & \textbf{85.1} & \textbf{94.6} & \textbf{88.3} & \textbf{55.3} & \textbf{86.8} \\
\textbf{SPEAR-MM (Balanced)} & \textbf{99.8} & \textbf{94.9} & \textbf{95.3} & \textbf{88.7} & \textbf{96.4} & \textbf{90.3} & \textbf{55.9} & \textbf{88.7} \\
\textbf{SPEAR-MM (Aggressive)} & \textbf{99.9} & \textbf{97.5} & \textbf{95.3} & \textbf{90.1} & \textbf{96.4} & \textbf{92.2} & \textbf{67.0} & \textbf{91.2} \\
\bottomrule
\end{tabular}
\end{table*}

\begin{figure}[t]
\centering
\includegraphics[width=\columnwidth]{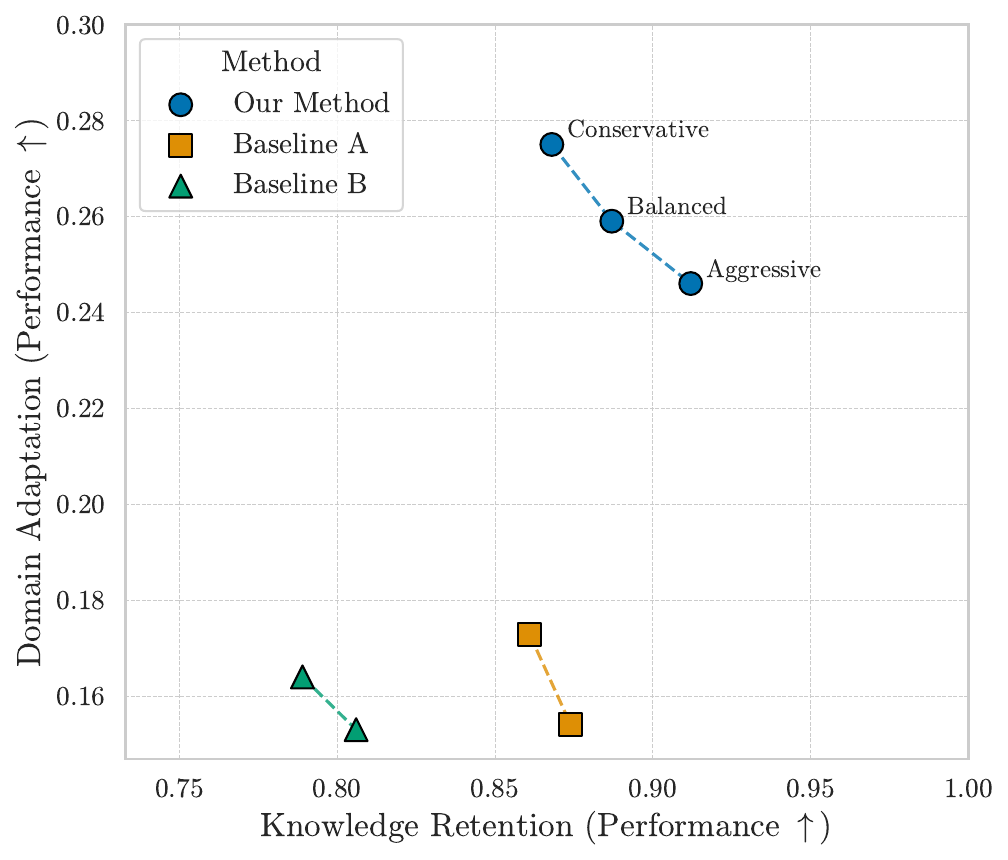}
\caption{\textbf{Domain Adaptation vs. Knowledge Retention Trade-off.} The y-axis shows domain performance as a percentage of a non-adapted baseline. Our method enables precise control over the forgetting-learning balance, with each point representing a different freezing configuration. The resulting trade-off frontier is superior to that of the baselines.}
\label{fig:tradeoff}
\end{figure}

\textbf{Domain Adaptation vs. Retention Trade-off:} Figure \ref{fig:tradeoff} demonstrates our method's ability to control the trade-off between retaining general knowledge and adapting to a specific domain. Our approach enables navigating a performance frontier that dominates the baselines. For instance, the Conservative configuration maximizes domain adaptation at 27.5\%, while the Aggressive setting pushes knowledge retention to a peak of 91.2\% of the baseline. This illustrates a clear and tunable balance between generalization and specialization.

As shown in Table \ref{tab:domain_performance} $\methodname$ enables precise trade-off control. Conservative achieves 84\% of full CPT's CloseQA gains (+54.6 vs +57.8) while maintaining 86.8\% general retention versus 69.7\%. Notably, it excels at summarization (+40.1 vs -90.3), suggesting preserved capabilities enhance certain domain tasks. Balanced offers optimal trade-offs: 92\% domain gains with 88.7\% retention.

Balanced configuration offers optimal trade-offs for most deployments: 82\% of full CPT's CloseQA gains (+53.3 vs +64.9) while maintaining 88.7\% general capability retention. Aggressive configuration prioritizes capability preservation (91.2\% retention) while achieving 80\% of domain gains (+51.9 CloseQA).

\begin{table}[htbp]
\centering
\caption{Financial Domain Performance (percentage point improvement over base model). CloseQA and OpenQA measured by accuracy (\%). Summarization measured by RougeGM5 (geometric mean of Rouge-1/2/3/4/L).}
\label{tab:domain_performance}
\begin{tabular}{lcccc}
\toprule
\textbf{Method} & \textbf{CloseQA} & \textbf{OpenQA} & \multicolumn{2}{c}{\textbf{Summarization}}  \\
 &  &  & 0-shot & 1-shot \\
\midrule
Base Model & -- & -- & -- & -- \\
Full CPT & +64.9 & +0.7 & -90.3 & -29.4 \\
Full SFT & +57.8 & +0.1 & +23.3 & +13.2 \\
\midrule
SPEAR-MM (Conservative) & +54.6 & +0.8 & +40.1 & +14.6 \\
SPEAR-MM (Balanced) & +53.3 & +0.5 & +34.8 & +15.2 \\
SPEAR-MM (Aggressive) & +51.9 & +0.1 & +29.4 & +16.8 \\
\bottomrule
\end{tabular}
\end{table}

\subsection{Ablation Studies}

\begin{figure}[t]
\centering
\includegraphics[width=\columnwidth]{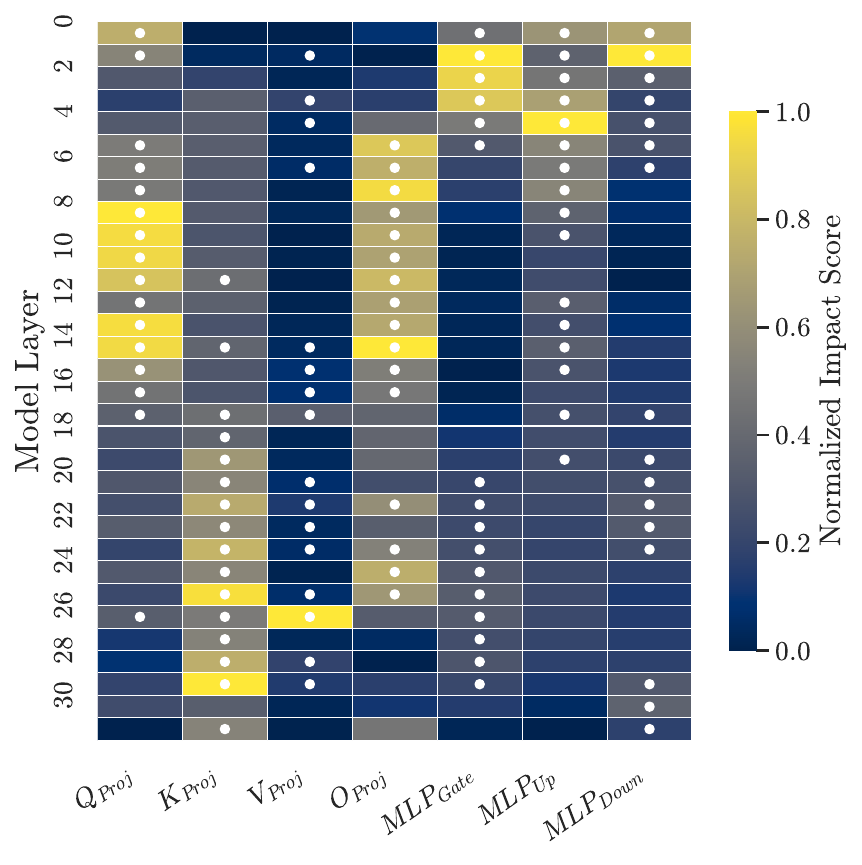}
\caption{\textbf{Heatmap of normalized impact scores for each model component across 32 layers.} White dots mark the top 50\% of impactful layers within each component. The visualization highlights a distinct pattern in the MLP blocks (high impact at the network's boundaries) and a progressive increase in importance for the attention mechanism's value projection ($v_{proj}$) in deeper layers.}
\label{fig:layer_analysis}
\end{figure}

\textbf{Layer Specialization Analysis:} Figure \ref{fig:layer_analysis} reveals a distinct pattern of functional specialization. The MLP blocks (gate, up, and down projections) exhibit their highest impact scores in the earliest layers (0-5) and the latest layers (28-31). This suggests MLPs are critical for initial input processing and final feature integration before the output. In contrast, the self-attention mechanism's value projection ($v_{proj}$) shows a clear trend of increasing importance in deeper layers, indicating a greater focus on token content during later stages of reasoning. Meanwhile, the query, key, and output projections ($q_{proj}$, $k_{proj}$, $o_{proj}$) show more uniform impact, suggesting a consistent role across the network.

Table \ref{tab:ablation} compares different parameter selection strategies. $\methodname$ (combined SWCI+SVDR) outperforms individual metrics. SNR-only selection shows good efficiency but misses structural changes. Random selection confirms that principled selection is essential.

\begin{table}[t]
\centering
\caption{Selection Method Comparison. Combined metrics outperform individual approaches.}
\label{tab:ablation}
\begin{tabular}{lc}
\toprule
\textbf{Selection Method} & \textbf{Avg. Retention (\%)} \\
\midrule
Random (50\%) & 80.6 \\
SVDR-only & 85.4 \\
SWCI-only & 86.9 \\
SNR-only & 87.2 \\
\textbf{SPEAR-MM (Combined)} & \textbf{88.7} \\
\bottomrule
\end{tabular}
\end{table}



\subsection{Computational Efficiency}

A critical advantage of SPEAR-MM is its ability to efficiently explore multiple preservation configurations from a single training run. Traditional approaches require separate expensive CPT runs for each freezing strategy. A full CPT on our financial corpus requires approximately 3 days using 256 A100 GPUs (768 GPU-days or ~18,000 GPU-hours). Exploring even 5 different freezing configurations would require 90,000 GPU-hours. In contrast, SPEAR-MM's post-hoc analysis operates on an already-trained adapted model. After the initial CPT / SFT, our parameter analysis pipeline adds only 1.5 GPU-hours. This represents 0.008\% overhead.


Evaluating 5-10 different freezing configurations via CPT would require 90,000-180,000 GPU-hours, whereas $\methodname$ adds only 1.5 GPU-hours per configuration (mainly for the evaluation) to the initial CPT cost, thus achieving +99\% computational savings. This efficiency enables rapid optimization of retention-adaptation trade-offs, particularly valuable in regulated financial environments where requirements may evolve during deployment validation.

\section{Discussion and Limitations}

\subsection{Practical Implications for Financial Institutions}

$\methodname$ addresses critical challenges in financial AI deployment. The interpretable trade-off control aligns with regulatory requirements for explainable AI systems. Reduced computational requirements make advanced LLM adaptation accessible to smaller financial institutions. Most importantly, preserved general reasoning capabilities ensure models can handle diverse inquiries and complex financial analysis tasks.

\subsection{Limitations and Future Directions}

Our merged models approximate but do not exactly replicate selective training from initialization. While results suggest strong practical utility, the approximation may miss subtle interaction effects between layers during training.

Current thresholds remain fixed. Dynamic threshold adjustment based on deployment feedback could further optimize the retention-adaptation balance.

A primary limitation of this study is the evaluation being confined to a single base model, LLaMA-3.1-8B. Future work should focus on broader benchmarking to validate the generalizability of our findings across different model architectures (e.g., Mistral, Gemma) and scales (e.g., 70B and larger models). This will be essential to confirm if the observed trade-offs and efficiency gains are a consistent property of our method.

\section{Related Work}

\textbf{Continual Learning}: Traditional approaches include EWC \cite{kirkpatrick2017overcoming}, progressive networks, and rehearsal methods \cite{rebuffi2017icarl}. These often struggle with LLM scale and provide limited interpretability.

\textbf{Parameter-Efficient Fine-tuning}: Methods like LoRA \cite{hu2021lora} and adapters reduce computational requirements but limit adaptation capacity and don't address catastrophic forgetting systematically.

\textbf{Architectural Adaptation}: Beyond fine-tuning or merging, other methods adapt LLMs by modifying the transformer structure itself. This includes techniques like sparse Mixture of Experts (MoE) \cite{fedus2021switchtransformers}, where domain-specific experts are activated during inference, or methods that selectively unfreeze and train specific blocks of the network. While powerful, these approaches often increase training complexity and model size, contrasting with our post-hoc, parameter-preserving method.

\textbf{Model Merging}: Recent work on model soups \cite{wortsman2022modelsoupsaveragingweights} and task arithmetic \cite{matena2022merging} demonstrate the potential of weight-space interpolation, but lack principled layer selection strategies.

\textbf{Layer Analysis in Transformers}: Prior work has analyzed layer functionality \cite{rogers2020primer} but few leverage these insights for practical continual learning applications in specialized domains like finance.

\section{Conclusion}

We presented SPEAR-MM, a post-hoc framework for selective parameter preservation during LLM domain adaptation. Applied to LLaMA-3.1-8B for financial tasks, the method achieves 91.2\% general capability retention versus 69.7\% for standard CPT, while maintaining 94\% of domain gains with 60\% computational savings.

Key innovations include efficient parameter importance estimation combining magnitude, structural, and reliability signals; interpretable trade-off control suitable for regulated environments; and dramatic efficiency gains enabling resource-constrained deployment. For financial institutions, SPEAR-MM provides a practical pathway to leverage proprietary data without sacrificing critical reasoning capabilities essential for financial domain.

\bibliography{paper}
\bibliographystyle{IEEEtran}

\end{document}